\documentclass[12pt,reqno,a4paper]{amsart}
\title[Image Registration]{MI image registration using prior knowledge}
\usepackage{graphics}
\usepackage{amsthm}
\usepackage{amsxtra}
\usepackage{amsfonts}
\usepackage{amssymb}
\usepackage{setspace}
\usepackage{epsfig}
\usepackage{graphicx}
\usepackage{amssymb}
\newcommand{\R}{\mathrm{I}\!\mathrm{R}}

\def\beq#1{
\begin{equation}\label{#1} 
}
\def\eeq{\end{equation}}
\def\vsep#1{{\vrule height #1 depth 0pt width 0 pt}}

\author{W. Jacquet \&  P. de Groen}

\newtheorem{the1}{Theorem}

\begin{document}

\maketitle
\begin{abstract}
Subtraction of aligned images is a means to assess changes in a wide variety of clinical applications.
In this paper we explore the information theoretical origin of Mutual Information (MI),
which is based on Shannon's entropy.
However, the interpretation of standard MI registration as a
communication channel suggests that MI is too restrictive a criterion.
In this paper the concept of Mutual Information (MI) is extended to (Normalized)
Focussed Mutual Information (FMI) to incorporate
prior knowledge to overcome some shortcomings of MI.
We use this to develop new methodologies to successfully address
specific registration problems, the follow-up of dental restorations,
cephalometry, and the monitoring of implants.


\end{abstract}

\newpage
\section{Introduction}
\setcounter{equation}{0}
In a wide variety of clinical applications subtraction of well-aligned
(or well-registered) images is a standard tool to monitor an evolution \cite{Maes2003}.
We study in this paper the popular registration criterion Mutual
Information (MI) \cite{Viola1996}, which is based on the joint entropy of images.
Shannon \cite{Shannon1948} introduced this type of entropy in the context of a communication channel
and Collignon \cite{Collignon1998} applied Shannon's model to MI-registration.
On the one hand, modeling image registration using Mutual
Information (MI) as a functional on a ``communication channel''
imposes unwanted restrictions in the context of image registration, and on the other hand,
the basic mutual information is not rich enough: spatial
information is lost, and it does not incorporate prior knowledge.
In traditional Mutual Information \cite{Viola1996} the gray value combinations
of all pixels in the overlap of reference image and floating test image are equally weighed.
In a probabilistic context this means that equal probability is attributed to each pixel in the overlap.
In \cite{Jacquet2007} we studied weighing and introduced Gauss Focussed Mutual Information (GFMI).
Here we place this in a probabilistic context,
and introduce mutual information with respect to a more general non-uniform distribution.
Prior knowledge related to a specific registration problem
is translated into a sampling distribution emphasizing the contribution
in the neighborhood of structures the practitioner wants to align or
structures that might contribute to the alignment,
and reduce the contribution of regions unimportant to the practitioner or unimportant to the alignment.
When monitoring implants or dental restorations, an obvious element
of prior knowledge is the radio opacity of these implants and restorations.
When 2D/3D images of a hollow longitudinal bone structure are to be
aligned it is natural to use edge detection to model the geometry of the bone.
This geometry can be brought into the alignment process as prior knowledge
through the sampling distribution.
In dental bitewing images used as a means to assess the evolution
of local phenomena, such as monitoring (small) dental lesions \cite{Jacquet2007},
it is the practitioner who has to provide the information about
which structure (tooth) that has to be aligned.
In Jacquet et al.\ \cite{Jacquet2007} Gaussian Focussed Mutual
Information (GFMI) is compared to a Region Of Interest (ROI) approach.
It was shown that the former approach in combination with registration based on affine transformations
is particularly well suited to align rigid parts when the context
of the underlying structure is relevant to the alignment.
When ROI is used without prior segmentation the resulting MI
(and therefore also the quality of the registration)
is highly dependent on the amount of background contained in the ROI.

In Section \ref{Sec:Image_Registration} image registration, the alignment of images, is formally defined.
Intrinsic registration methods are introduced in Section \ref{Sec:Intrinsic_Methods},
joint entropy of images in Section \ref{Sec:Joint_entropy}.
Information theory \cite{Shannon1948} is briefly presented in Section \ref{Sec:Information_Theory}.
In Section \ref{Sec:Focussed_MI} mutual information
based registration is placed in this information theoretical context,
and extended to incorporate prior knowledge.
In Section \ref{Sec:Applications} we use this extension to
develop new methodologies to successfully address specific
registration problems, the follow-up of dental restorations,
cephalometry, and the monitoring of mandibular implants.
The same ideas can be used for registration of 3D images; currently we are developing software
and test strategies for hip-, knee-, and shoulder implants.
We do not address issues of medical interpretation and diagnosis.

\section{Image registration}\label{Sec:Image_Registration}
\setcounter{equation}{0}
In practice, a gray scale image is a rectangular array of pixels
and a function $u$ assigning one of $K$ gray value bins $\{\,1\,,\,\cdots\,,\,K\,\}$ to each pixel.
It can be considered as the discretization of a continuous
function on a subset $A$ of $\R^2$, with values in the interval $[0,1]$.
Let $u$ and $v$ be two (continuous) images with domains $A$ and $B$ $\subset \R^2$ respectively.
Let $\mathcal{A}$ be a class of smooth bijective mappings from $\R^2$ to $\R^2$.
The idea of image registration is to find a transformation
$T \in \mathcal{A}$ such that image $u$ and
the transformed image $v_T \, := \, v \circ T^{-1}$ are as similar as possible.
Given a measure $I$ of similarity between images, optimal
registration can be defined as the problem to find
$$\arg \, \max_{T \in \mathcal{A}} I(u,v_T) \textrm{,}$$
the domain-transformations maximizing this measure.
In this setting we will refer to $u$ as the reference image,
$v$ as the test image, and $v_T$ the floating image.
Different classes of transformations are required depending on the registration applications.
\section{Intrinsic registration methods}\label{Sec:Intrinsic_Methods}
\setcounter{equation}{0}
In Maintz and Viergever \cite{Maintz1998} a classification of registration methods is introduced.
They call a method ``intrinsic'' when it relies only on patient generated image content,
and ``extrinsic'' when objects foreign to the patient are introduced into the scene
of which an image is taken to serve as reference to the alignment process.
The intrinsic methods are split into landmark based,
segmentation based, and voxel/pixel property based registration methods.
In landmark based and segmentation based registration corresponding
structures are indicated or extracted from
reference and test image, to be used pairwise as input for the alignment procedure.
A voxel/pixel property based registration criterion is a criterion directly
linked to the discrete two-dimensional gray value maps (\ref{Gray}).

\begin{figure}[htbp]
\begin{center}
\input{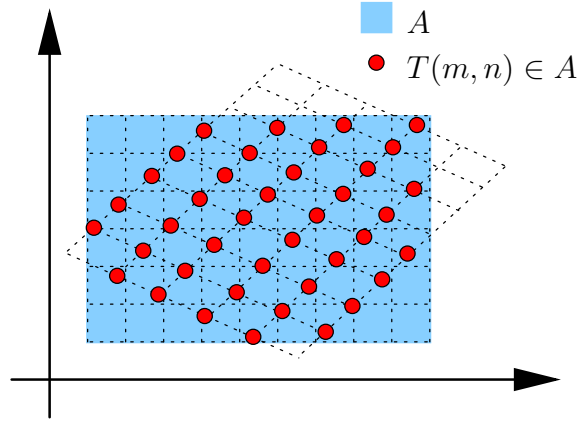}
\end{center}
\caption{Mapping of the test image pixels on the reference image through
s\mbox{transform $T$} cf. \cite{Jacquet2007}.\label{fig:figuurtje}}
\end{figure}

To be precise, in the process of registration we consider two discrete images,
denoted by $u$ and $v$, and transformations $T$, and the two-dimensional gray value maps:
\beq{Gray}
\begin{array}{c}
\widetilde{\mathbf{g}}_{T}(m,n) \, := \,(\widehat{u}(T(m,n)),v(m,n)) \textrm{ ,}
\end{array}
\eeq
where $\widehat{u}$ is the gray value bin of an interpolation of
the known values of $u$ at $T(m,n)$ within $A$ (\mbox{Fig. \ref{fig:figuurtje}}).
Several interpolating techniques can be applied.
We will adopt bilinear interpolation.

\section{Joint entropy of image pairs}\label{Sec:Joint_entropy}
\setcounter{equation}{0}

Consider the function that counts the number of times a gray value combination occurs
in the intersection of the reference image with the transformed test image:
\beq{Count}
C_{T}(k,\ell)  := \sharp \, \{(m,n)\, | \, T(m,n)\in A \, \wedge\,
(k,\ell)=(\widehat{u}(T(m,n)),v(m,n))\,\}
\end{equation}
If test and reference image are equal and well aligned with respect to each other,
the frequency of unequal pairs of pixel values or off-diagonal elements is zero.
If the well-aligned images slightly differ, off-diagonal elements may become non-zero but are expected to be small.
This suggests that a measure based on the function $C_{T}$ can be the basis for aligning images.
We divide $C_{T}$ by the total number of pixels in the intersection of reference and transformed test image
to eliminate the dependence on the image size.
The result is a two-dimensional probability distribution, called the joint probability of the images.
Let $p_{k\,\ell}$ denote the probability linked to the combination of gray value classes $k$ and $l$,
and let $p_{k\,\bullet}$ and $p_{\bullet \, \ell}$ denote its marginal distributions,
\beq{PropH1}
\begin{array}{llr}
p_{k \, \ell}      &:= \frac{C_{T}(k,\ell)}{\sum_{ij}C_{T}(i,j)}&\textrm{ joint probability,}\\
p_{k\,\bullet}     &:= \sum_l \, p_{k\,\ell}                    &\textrm{ marginal linked to the reference image,}\\
p_{\bullet \, \ell}&:= \sum_k \, p_{k\,\ell}                    &\textrm{ marginal linked to the test image.}
\end{array}
\end{equation}
Although $p_{k \, \ell}$ has all formal aspects of a probability,
it is merely a relative frequency not linked to a stochastic experiment at present.

\subsection*{Shannon Entropy}

A finite probability model is characterized by a set of $n$ elements (outcomes)
and their probabilities $p_1,p_2, \ldots , p_n$.
Shannon defines a measure of uncertainty $H$ on the class of all finite probability models,
$\{(p_1,p_2, \ldots , p_n)\,|\, n \in \mathbf{N} \textrm{ and } \sum p_i = 1 \, \}$, satisfying:
\beq{PropH}
\begin{array}{lcl}
\multicolumn{3}{l}{H(p_1,p_2, \cdots , p_n) \textrm{ is a non-negative continuous function of the $p_i$,}}\\
\multicolumn{3}{l}{H_n := H(1/n,1/n, \cdots , 1/n)  \textrm{ is an increasing function of $n$,}}\\
H(p_1,\ldots,p_{n-1},p_n) & = &H(p_1,\cdots,p_{n-1}+p_n) \\
                          &   &+ (p_{n-1}+p_n) H(\frac{p_{n-1}}{p_{n-1}+p_n},\frac{p_n}{p_{n-1}+p_n})\textrm{ .}
\end{array}
\end{equation}
Implicitly, but clear from his application:
\beq{PropH2}
H(p_1,p_2, \ldots , p_n) = H(p_{\sigma(1)},p_{\sigma(2)}, \ldots , p_{\sigma(n)}) \textrm{ ,}
\end{equation}
with $\sigma$ a permutation of the indices $1 \cdots n$.

\begin{the1} If $H$ satisfies the above requirements then:
\beq{Shannon}
 H(p_1,p_2, \ldots , p_n) \, = \, - K \sum_{i=1}^{n} \, p_i\, \ln p_i \textrm{ ,}
\end{equation}
where $K$ is a positive constant.
\end{the1}
The proof of this theorem can be found in \cite{Shannon1948}.
For $K = 1$ is $H$ referred to as the Shannon entropy.
The Shannon entropy measures uncertainty about a precise outcome of an experiment linked to a probability distribution.

\subsection*{Joint entropy}
The joint entropy $H(u,v_T)$ of two images $u$ and $v_T$ is the Shannon entropy
of the joint probability $\{p_{k\,\ell}\}$ of the images.
The entropy $H(u)$ of the reference image is the Shannon entropy of $\{p_{k\,\bullet}\}$ ,
the entropy $H(v_T)$ is the Shannon entropy of $\{p_{\bullet\,\ell}\} \,$.

When two images are well aligned but intensities of the same structure differ, the probabilities will still concentrate
and the Shannon entropy will not be affected since it is invariant under permutation of elements -- see (\ref{PropH2}).
Therefore a similarity measure based on the entropy can be applied when different image acquisition modalities are used.
This is in addition to the absence of preprocessing, a reason why Entropy based similarity measures are popular in
multi-modal registration applications at present.

\section{Information theory applied to image registration}\label{Sec:Information_Theory}

\begin{figure}
\begin{center}
\includegraphics{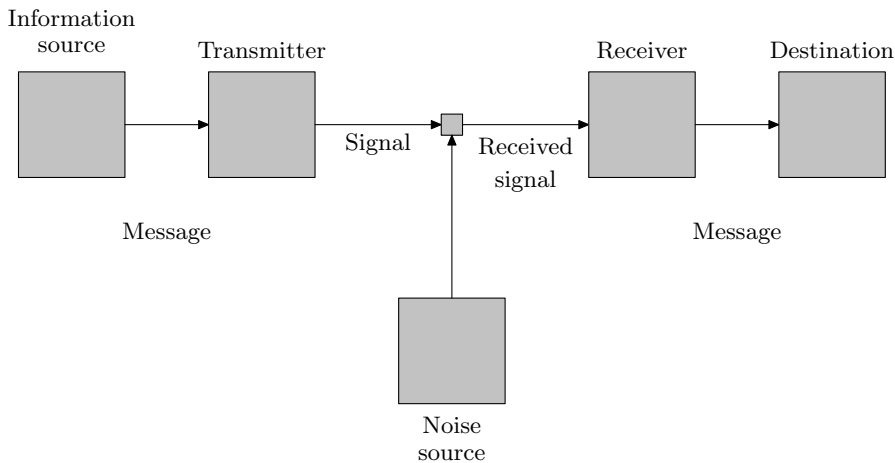}
\end{center}
\caption{Schematic diagram of a general communication system \cite{Shannon1948}}
\label{ComSys}
\end{figure}

The mathematical theory of communication developed by Shannon \cite{Shannon1948} is based on a general communication system
described by Fig. \ref{ComSys}.
Shannon attempts to build an indicator for the quality of the combination transmitter, channel and receiver.
A communication system should be designed to enable passing all acceptable messages generated by the information source.
When a language such as English is used by the information source as coding system,
not all choices of acceptable messages are equally likely to be generated.
E.g.\ the occurrence of the sequence of symbols ``My name is'' is more likely than the sequence ``Prescience is''.
Shannon introduces an artificial information source consisting of a Markov process as a model for the English language.
The introduction of the surrogate source allows a study of the quality and limitations of a communication system
with probabilistic methods.
The English language is not the most efficient coding system. 
To study a communication system in combination with an English information source,
it is important to define a measure of information content of a phrase or of its complement -- the redundancy.
The Shannon Entropy $H$ is build to measure how uncertain we are about a specific phrase (outcome), or the complement,
how probable the specific phrase is when we suppose it has been generated by a Markov process modeling the English language.

Let us try to understand the requirements that define $H$.
The first requirement is continuity: there is no clear reason to introduce ``jumps''.
The continuity requirement does not seem to be too restrictive.
The second requirement states that if the number of possible outcomes increases,
and if all outcomes are equally probable, the uncertainty about the realization of one particular outcome increases.
The last requirement can be understood from modeling of a language at different levels.
E.g.\ modeling the language at the level of characters should be compatible with, but less distinctive than,
modeling the language at the level of words.
To be precise: when a model is refined, the entropy should grow.
If a system is split into two sub-systems,
the increment of the entropy is equal to the probability of the systems multiplied by the entropy within the system.

Collignon \cite{Collignon1998} applied the model of a communication channel to image registration under the following assumptions:
\begin{itemize}
\item Consider the test image to be the transmitted signal.
\item Take the reference image to be the received signal.
\item The communication channel is determined by the registration parameters.
\item Optimizing the mutual information between the signals is equivalent to
the design of an optimal communication channel.
\item Both images are assumed to represent the same scene,
and their multi-modal differences are considered a noise generated by the communication channel.
\end{itemize}

Inspecting the construction of the joint probability of two images in more detail,
one can see that it is induced by the
gray value mapping $ \widetilde{\mathbf{g}}_{T}$ transforming the uniform distribution on the pixels $T(m,n)\in A$
from the test image into an induced probability on $[1\,,\,\cdots\,,\,K]\times[1\,,\,\cdots\,,\,K]$.
Therefore a process generating the joint probability consists of the sequential uniform random pick of nodes $T(m,n)$
located within the domain of the reference image $A$ and the evaluation of $\widehat{u}(T(m,n))$ and $v(m,n)$.
The signal becomes the evaluations $u_t:=\widehat{u}(T(m_t,n_t))$, the channel is defined through the
conditional probabilities $ p_{k\,u_t}/p_{\bullet \, u_t}$ where $k = 1, \cdots , K$.
The origin of the dispersion of these conditional probabilities is determined by the modalities used
to acquire the images, the instrument settings, and the quality of the alignment.
The joint probability is only a first order model
in the sense that all information concerning the location of the gray value couples in both images is ignored.
This is comparable to the generation of a sequence of letters and spaces with the frequency of occurrence in the
English language and its use as a model for an English phrase.
Spatial relation of pixels corresponds to the sequential order of the characters in a sentence.
This spatial relation is discarted altogether in MI.

Although the effort to translate the image registration problem into a problem about
modeling a communication channel is appealing from a theoretical validation viewpoint
it limits the search of new criteria,
and it directs the search towards the statistical estimation of probability densities
naturally emerging from an unknown ``communication system''.

The third condition of (\ref{PropH}) does not seem mandatory in a image modeling context.
Therefore, the substitution of e.g.\ the logarithm by a square root
or any other convex function in (\ref{Shannon}) is an interesting line of thought,
influencing the robustness of the criterion.
Hughes and Daubechies   \cite{Hughes2007} propose simpler alternative metrics
based on the joint probability of overlapping images.
The introduction of the spatial relation can be achieved by taking into account neighboring gray value couples.
This approach has been explored a.o.\ in \cite{Rueckert2000};
in  \cite{Pluim2000b} and \cite{Guo2006} gradient information is calculated and incorporated into the functional,
and in \cite{Rodriguez1998} gray value differences of neighboring pixels/voxels are used.
The introduction of spatial information in MI can also be achieved
through blurring of the images before registration.
However, this blurring may cause loss of a significant amount of information.

Another line of thought, apart from spatial considerations and robustness,
is to incorporate prior knowledge about scene and application by
replacing the uniform sampling by a sampling according to a suitable distribution.
A higher probability can be attributed to regions required to align particularly well,
or to structures relevant to the alignment.
In this paper we shall consider non-uniform distributions,
and introduce different methodologies to incorporate prior knowledge into the sampling distribution
tailored to specific applications.

Another issue in registration based on pixel/voxel based criteria is the image overlap \cite{Studholme1999}.
Size and content of the overlap may change considerably during the registration.
When dealing with images with very high aspect ratio even "small" transformations can have an important influence on the overlap.
To reduce the sensitivity of MI to these changes in overlap statistics, and to minimize the resulting misalignments,
the Normalized Mutual Information $Y$ \cite{Studholme1999}
and the Entropy Correlation Coefficient (ECC) \cite{Maes1998} have been introduced.
These criteria have shown a better robustness to changes in overlap statistics than MI does.
Therefore, we will adapt $Y$ to non-uniform distributions and use it in our applications.
We will not use an analogous adaptation of the ECC, since ECC is directly related to $Y$.

\section{(Normalized) Focussed Mutual Information with respect to a density}\label{Sec:Focussed_MI}

Shannon \cite{Shannon1948} extends the entropy of a finite probability distribution
to the entropy of a probability distribution with density $f$ on a domain $\Omega$:
\beq{Entropy}
\begin{array}{c}
H(f) \, = \displaystyle \, -\int_{\Omega} \, f \, log(f) \mbox{ . }
\end{array}
\end{equation}
Let $u$ and $w$ be continuous images with intersecting domains $A$ and $B$, $f$ a probability density function on $A \cap B$
and $\mathbf{g}$ the function assigning to each point $\omega$ in the intersection $A \cap B$
the couple of gray values $\mathbf{g}(\omega):=(u(\omega),w(\omega))$.
Denote by $f_{\mathbf{g}}$ the probability density function on $[0,1]\times[0,1]$, generated through $\mathbf{g}$,
and denote by $f_u$ and $f_w$ the probability density functions on $[0,1]$ generated by $u$ and $w$, respectively.
Note that $f_u$ and $f_w$ are the marginal distributions of $f_{\mathbf{g}}$ with respect to $y$ and $x$ respectively.
We define
\begin{itemize}
\item
{\em Focussed Mutual Information}
of the images $u$ and $w$ with respect to a density $f$ on the overlap of the images $A \cap B$ as follows:
\beq{MIcont}
                  MI_f(u,w) :=    \,  H(f_u) \, + \, H(f_w) \, - \, H(f_{\mathbf{g}}) \, \mbox{, }
\end{equation}
\item
{\em Normalized Focussed Mutual Information}
of the images $u$ and $w$ with respect to a density $f$ on the overlap of the images $A \cap B$ as follows:
\beq{Y_f}
 Y_f(u,w)  := \frac{H(f_u)+H(f_w)}{H(f_{\mathbf{g}})} \, \mbox{, }
\end{equation}
\item
{\em Focussed Entropy Correlation Coefficient}
of the images $u$ and $w$ with respect to a density $f$ on the overlap of the images $A \cap B$ as follows:
\beq{ECC}
ECC_f(u,w) := \frac{2 \, MI_f(u,w)}{H(f_u)+H(f_w)} \, \mbox{. }
\end{equation}
\end{itemize}

These generalizations return to the original concepts if $f$ is chosen as the uniform distribution,
i.e. if there is no focussing.
So Focussed Mutual Information see Eq. (\ref{MIcont}) returns to MI introduced by Collignon \cite{Collignon1998} and Maes \cite{Maes1998},
Normalized Focussed Mutual Information see Eq.\ (\ref{Y_f}) returns to $Y$ introduced by Studholme \cite{Studholme1999}, and
and the Focussed Entropy Correlation Coefficient see Eq.\ (\ref{ECC}) returns to ECC of  Maes \cite{Maes1998}.

The Focussed Mutual Information with respect to a density function $f$ is an extension of the Gauss Focussed Mutual Information
introduced by Jacquet et al.\ \cite{Jacquet2007}.
More precisely, given two continuous images $u$ and $w$ with domains $A$ and $B$ respectively,
and $f = \sum a_i \, f_i$ the  convex combination of a finite number of normal density functions $f_i$,
normalized to be a probability density function on $A \cap B$,
then $MI_f(v,w)$ is the Gauss Focussed Mutual Information.

\subsection*{Approximation of $MI_f$, $Y_f$, and $ECC_f$}

We consider two discrete images, denoted by $u$ and $v$.
Let $(k,\ell)$ be a gray value combination.
Denote by $D(k,\ell)$ the set of all pixel pairs in the intersection that have (approximately)
the gray value combination $(k,\ell)$,
 $$D(k,\ell) \, := \,\{(m,n)| T(m,n)\in A \, \wedge\,(k,\ell)=(\widehat{u}(T(m,n)),v(m,n))\,\}\textrm{ . }$$
As before in (\ref{Count}), we define $C_T(k,\ell)$
as the number of elements of $D(k,\ell)$ in which each element contributes equally (or is equally weighed).
We introduce $W_T(k,\ell)$ as a weighed sum in which the contribution of each pixel pair is
proportional to the focus probability assigned to its location, and we normalize those quantities into probabilities:
\beq{probabilities}
\begin{array}{rcccrcc}
C_{T}(k,\ell)        & := & \displaystyle \sum_{(m,n)\in D(k,\ell)} \, 1 \,\,\,          & \textrm{ ,}&
W_{T}(k,\ell)        & := & \displaystyle \sum_{(m,n)\in D(k,\ell)} \, f(m,n)              \textrm{ ,}            \\
\\
p_{k\,\ell}          & := & \displaystyle \frac{C_{T}(k,\ell)}{\sum_{ij}C_{T}(i,j)}&  \textrm{ ,}&
\pi_{k\,\ell}        & := & \displaystyle \frac{W_{T}(k,\ell)}{\sum_{ij}W_{T}(i,j)}  \textrm{ ,}            \\
p_{k\,\bullet}       & := & \displaystyle \sum_l \,  p_{k\,\ell}                   &  \textrm{ ,}&
\pi_{k\,\bullet}     & := & \displaystyle \sum_l \, \pi_{k\,\ell}                       \textrm{ ,}            \\
p_{\bullet \, \ell}  & := & \displaystyle \sum_k \,  p_{k\,\ell}                   &  \textrm{ ,}&
\pi_{\bullet \, \ell}& := & \displaystyle \sum_k \, \pi_{k\,\ell}                    \textrm{ .}
\end{array}
\end{equation}
Here $p_{k\,\ell}$ is the joint probability of the images, see Eq.\ (\ref{PropH1}),
$\pi_{k\,\ell}$ the focussed joint probability of the images,
$\pi_{k  \,\bullet}$ the focussed marginal linked to the reference image, and
$\pi_{\bullet \, \ell}$ the focussed marginal linked to the test image.
The corresponding (``focussed'') entropies are:
\beq{entropies}
\begin{array}{lcl}
\widehat{H}_f(u,v_T)\displaystyle   &=& -\sum_{k\,\ell} \, \pi_{k\,\ell}  \,\log(\pi_{k\,\ell})\, \mbox{,}\\
\widehat{H}_f(u)    \displaystyle   &=& -\sum_{k} \, \pi_{k\,\bullet}  \,\log(\pi_{k\,\bullet})\, \mbox{,}\\
\widehat{H}_f(v_T)  \displaystyle   &=& -\sum_{\ell} \, \pi_{\bullet \,\ell}  \,\log(\pi_{\bullet \, \ell})\, \mbox{. }\\
\end{array}
\end{equation}
We introduce approximations of $MI_f(u,v_T)$, $Y_f(u,v_T)$ and $ECC_f(u,v_T)$:
\beq{MI}
\begin{array}{lcl}
\widehat{MI}_f(u,v_T)  \ &=& \widehat{H}_f(u) + \widehat{H}_f(v_T) - \widehat{H}_f(u,v_T)\, \mbox{,}\\
\widehat{Y}_f(u,v_T) \vsep{2.5em}   &=& \displaystyle \frac{\widehat{H}_f(u) +
\widehat{H}_f(v_T)}{\widehat{H}_f(u,v_T)} \, \mbox{,}\\
\widehat{ECC}_f(u,v_T) \vsep{2.5em} &=& \displaystyle \frac{2\, \widehat{MI}_f(u,v_T)}
{\widehat{H}_f(u) + \widehat{H}_f(v_T)} \, \mbox{.}
\end{array}
\end{equation}

\section{Applications}\label{Sec:Applications}
In this section, we will introduce methodologies involving FMI and Digital Subtraction Radiography (DSR), tailored to specific clinical applications.
Each of the proposed registration methods will be a hybrid form between a landmark/segmentation and a pixel/voxel based method.
Anatomical structures, present in reference and test image, will be used to define a probability distribution $f$ on the reference image
incorporating the prior knowledge of the problem.
The trace distributions $f_T$ of the probability distribution $f$ on the intersection of the domains of reference image
and floating test image constitute the basis for a pixel/voxel based registration.
The registration criterion is the Normalized Focussed Mutual Information $Y_f$ see Eq.\ (\ref{Y_f}).

In a standard feature based registration, landmarks have to be identified both in reference image and test image.
In contrast, the focus distribution has to be defined for the reference image only and it
eliminates the need for accurate landmark detection and for pairwise landmark matching.
The locations of the landmarks are used only to define the regions of high probability; extreme accuracy is not needed.
Nevertheless it may be handy to add some landmarks to the test image (not necessarily accurately located),
to obtain an initial guess $T'$.
The search space of parameters of the transformation $T$  is limited to a region located symmetrically around the parameter values of $T'$.
In our experiments we will restrict to affine transformations as first order approximations.

\subsection{Dentistry and orthodontics}\label{subsec:Dentistry}
\begin{figure}
\begin{center}
\includegraphics*[width=7.5cm]{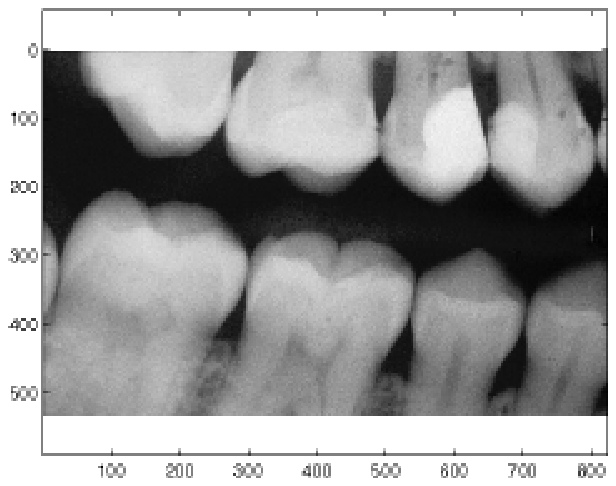}\hspace{2em}
\includegraphics*[width=7.5cm]{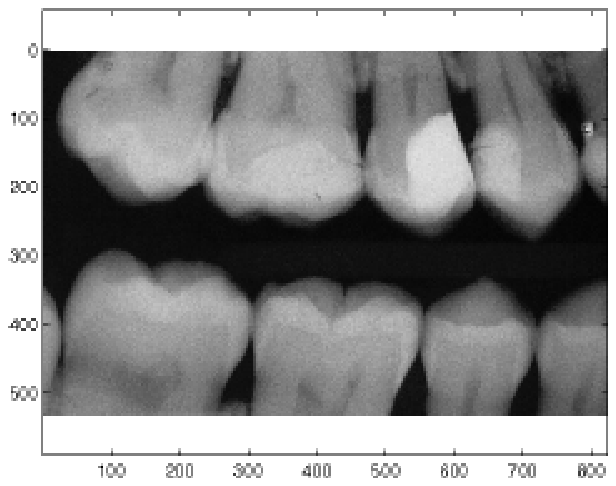}
\end{center}
\caption{\small\it Reference image (left) and Test image (right). \label{DENT_Ref_Test}}
\end{figure}

\setcounter{equation}{0}
The detection and evolution of periodontal diseases and of alveolar bone changes can be facilitated through
intra-oral radiography in combination with Digital Subtraction Radiography (DSR) \cite{Mol2004}.
Non intervention therapy, observation in combination with adequate preventive measures \cite{Kidd2003},
and therapies such as pulpa capping require the possibility to assess evolution.
DSR is a promising technique for the follow-up of small lesions \cite{HaiterHero2005}, restorations,
and pulpa capping.
The most common imaging techniques in dentistry and orthodontics are 2D.
The variation in geometry between image acquisitions most often results in irreversible distortions from one image to the other \cite{Webber1984}.
Satisfactory alignment of the whole image can only be obtained if the variation in geometry is sufficiently small.
The variation can only be small enough if the X-rayed scene can be considered rigid.
In dental applications this is most often not the case, due to natural tooth mobility,
orthodontic treatment or movement of the lower jaw with respect to the upper jaw.
Focussing the structures around the local phenomenon under study will allow for a better local alignment.
Such a structure can be indicated manually or by means of an automatic or semi-automatic procedure.
For the follow-up of small lesions Jacquet et al.\ \cite{Jacquet2007}
explore the use of FMI based on a convex combination of Gaussian distributions in combination with DSR.
In this study the marking of the tooth to be monitored is done in the reference image, manually with a digital tool.
In what follows we will give a case study of \mbox{(semi-)} automatic generation of a focus distribution
for monitoring the quality of a dental restoration over time.
In \mbox{Fig. \ref{DENT_Ref_Test}} we see two images of the teeth of one person, taken two years apart.
We want to register the restoration on the smaller upper molar.
To this aim we construct a focus distribution around the restoration in the reference image exploiting prior knowledge,
that restorations (and implants) are more radio-opaque than the surrounding natural material.
\par
\begin{figure}
\begin{center}
\includegraphics*[width=7.5cm]{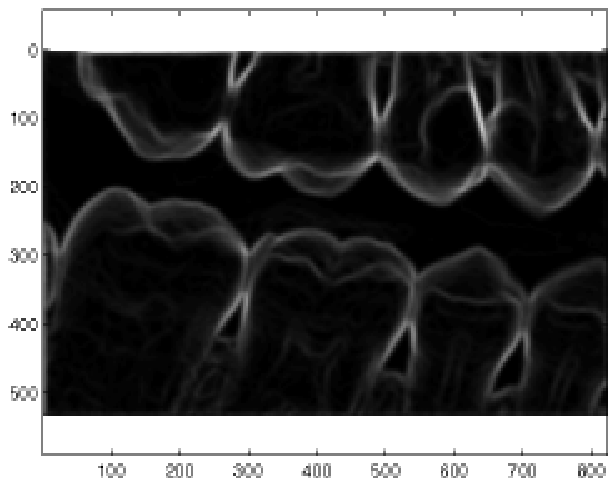}\hspace{2em}
\includegraphics*[width=7.5cm]{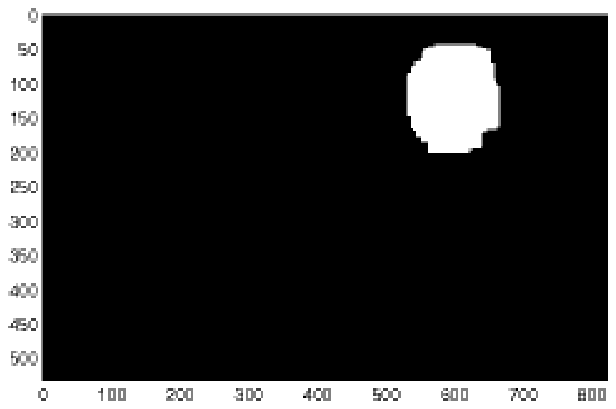}
\end{center}
\caption{\small\it Edge image (left) and closed and dilated restoration (right). \label{DENT_Edge_Resto}}
\end{figure}

\noindent{\bf Algorithm}
Automatic generation of a focus distribution aiming at the correct registration of consecutive images of a tooth restoration:
\begin{list}{(\arabic{enumiii}).}{\leftmargin3em \usecounter{enumiii}}
\item
Find (all) edges in the reference image by:
\begin{itemize}
\item median filtering to eliminate ``pepper and salt'' noise from the reference image.
\item computation of the modulus of the gradient.
\item convolution with a Gaussian kernel.
\end{itemize}
This results in \mbox{Fig. \ref{DENT_Edge_Resto} left}.
\item
Find a patch that contains the whole restoration:
\begin{itemize}
\item segmentation using a threshold to select the restoration.
\item morphological closing and dilation.
\end{itemize}
The indicator function of the patch is shown in \mbox{Fig. \ref{DENT_Edge_Resto} right}.
\item
Create the focus distribution:
\begin{itemize}
\item multiply the patch from step (2) and edge distribution produced in step (1).
\end{itemize}
\end{list}

\begin{figure}
\begin{center}
\includegraphics*[width=7.5cm]{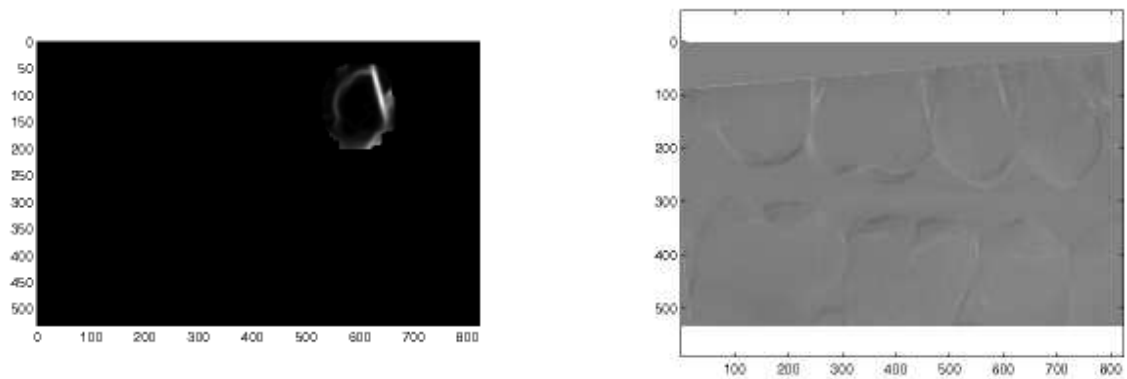}\hspace{2em}
\includegraphics*[width=7.5cm]{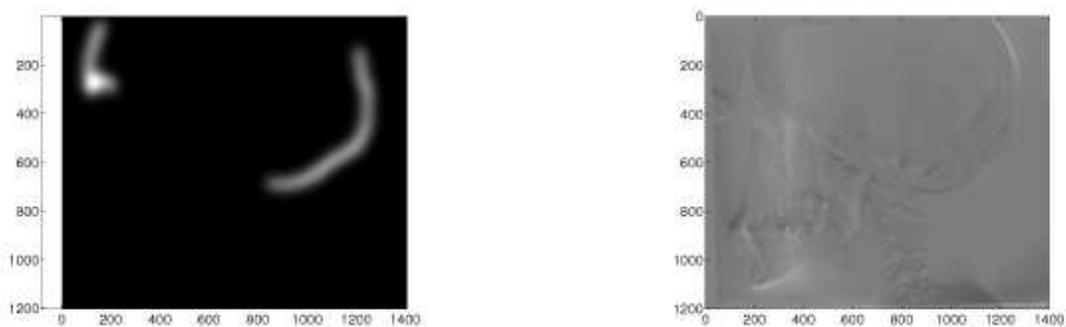}
\end{center}
\caption{\small\it Focus distribution $f$ (left) and subtraction image (right). \label{DENT_Foc_Diff}}
\end{figure}

FMI registration using this focus distribution results in \mbox{Fig. \ref{DENT_Foc_Diff} right}, showing a well aligned restoration.
One can think of first creating the patch selecting a part of the image containing the restoration,
followed by edge detection and convolution.
Working in this order we may easily create spurious edges due to the border of the indicator of the patch.

\subsection{Cephalometry}\label{subsec:Cephalometry}
\setcounter{equation}{0}
Cephalometry is used as a diagnostic tool and as a basis for treatment planning, but also to monitor and evaluate treatment results \cite{Ayoub2000}.
In clinical practice the evolution is assessed by superimposing consecutive lateral radiographs based on anatomical landmarks.

\begin{figure}
\begin{center}
\includegraphics*[width=7.5cm]{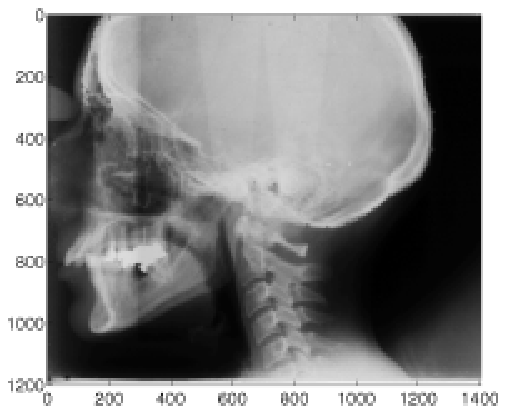}\hspace{2em}
\includegraphics*[width=7.5cm]{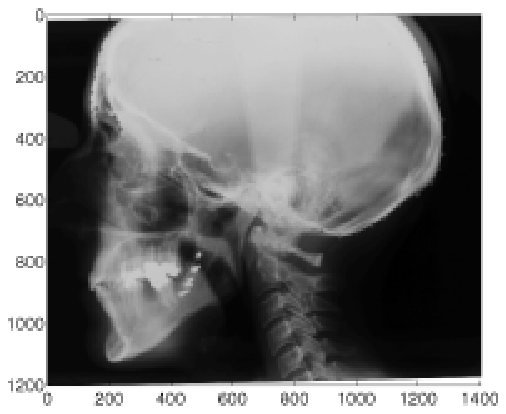}
\end{center}
\caption{\small\it Reference image (left) and Test image (right) with indicated features. \label{25RSP_Ref_Test_Points}}
\end{figure}

As a case study we applied FMI registration to an example of false maxillary prognathism.
A lack of growth of the mandible is corrected by means of a combined surgical and orthodontic treatment,
where the mandibular has been advanced.
A lateral radiograph is taken before treatment (\mbox{Fig.\ \ref{25RSP_Ref_Test_Points} left}),
and a follow up lateral radiograph is taken two years after treatment (\mbox{Fig.\ \ref{25RSP_Ref_Test_Points} right}).
The purpose of the images is the evaluation of skeletal stability,
and orthodontic treatment.

\begin{figure}
\begin{center}
\includegraphics*[width=7.5cm]{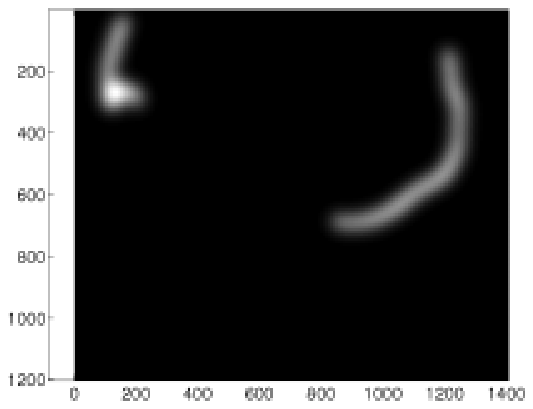}\hspace{2em} %
\includegraphics*[width=7.5cm]{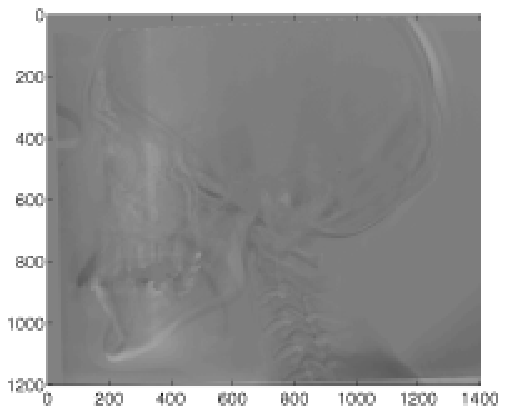}
\end{center}
\caption{\small\it Focus distribution $f$ (left) and subtraction image (right). \label{25RSP_Foc_Diff}}
\end{figure}

The practitioner is asked to indicate each structure to be used in the alignment procedure with a limited number of points (15)
in the reference image (\mbox{Fig.\ \ref{25RSP_Ref_Test_Points} left}) and the test image (\mbox{Fig.\ \ref{25RSP_Ref_Test_Points} right}).
The points in the reference image are transformed into a model of the skull using B-splines.
The resulting image is convolved with a Gaussian kernel, and used as distribution in an FMI registration -- see \mbox{ Fig.\ \ref{25RSP_Foc_Diff} left}.
The subtraction image shows clearly the effect of the orthodontic treatment,
and the extent of the surgical correction -- see \mbox{ Fig.\ \ref{25RSP_Foc_Diff} right}.

In the aligning process of the lateral radiographs of the skull the input of the practitioner can easily be reduced or removed.
The detection of the edges delineating the front and back of the skull can be fully automated and used as the input for the
FMI registration of the lateral radiographs.
Another line of thought is to use automatically detected landmarks in the reference image as prior knowledge to construct a focus distribution.
The automatic detection of cephalometric anatomical landmarks is promising e.g.\ \cite{Ciesielski2003} and \cite{Rudolph1998}.
In combination with the reduced need for accuracy of the localization of landmarks in a FMI they can provide the basis for a successful automated FMI registration algorithm.

An even more challenging application is the use of registration of lateral images of the skull in treatment planning.
Crucial in the decision to start the orthodontic and/or operative treatment of an adolescent is the detection of the end-of-puberty growth sprint.
For characterizing the growth curve we plan to study the evolution of the registration parameters,
more precise, the scaling needed to adjust consecutive images of the skull.

\subsection{Follow up of implants}\label{subsec:Implants}
\setcounter{equation}{0}
Digital subtraction of scans taken at time intervals can be used to monitor
the evolution of a prosthesis with respect to its wear and anchoring in the skeleton,
more particular to assess the relative movement of a prosthesis with respect to the cavity in which it resides.
The use of digital subtraction could be relevant to the detection and follow up of aseptic loosening of implants.
Focussing on the bone structure we can model the bone in its related surrounding soft tissue,
moreover we can eliminate the effect of the implant from the registration procedure,
using the hollow structure of the bones and the radio opacity of the implants as prior knowledge.

\par\noindent{\bf Algorithm}
Automatic generation of a focus distribution aiming at the correct registration of the bone structure surrounding an implant:
\begin{list}{(\arabic{enumiii}).}{\leftmargin3em \usecounter{enumiii}}
\item
Find (all) edges in the reference image by:
\begin{itemize}
\item median filtering to eliminate ``pepper and salt'' noise from the reference image.
\item computation of the modulus of the gradient.
\item convolution with a Gaussian kernel.
\end{itemize}
This results in an edge distribution focussing all the edges.
\item
Find the complement of a patch covering the implant:
\begin{itemize}
\item segmentation using a threshold to select the implant.
\item morphological closing and dilation.
\item creation of an indicator of the complement of the patch covering the implant.
\end{itemize}
\item
Create the focus distribution:
\begin{itemize}
\item multiply the patch from step (2) and edge distribution produced in step (1).
\end{itemize}
\end{list}
Only edges corresponding to structures not related to the implant will contribute to the FMI registration.
The reason to focus on the bone structure is that it becomes easy to measure the movement of the implants when the bone structure is well aligned.
In the case of dental implants the opposite procedure is more appropriate.
It is better to register the implant and evaluate the evolution of the surrounding bone tissue.
3D-2D projections will make displacement measurements unreliable.

\par
\begin{figure}
\begin{center}
\includegraphics*[width=7.5cm]{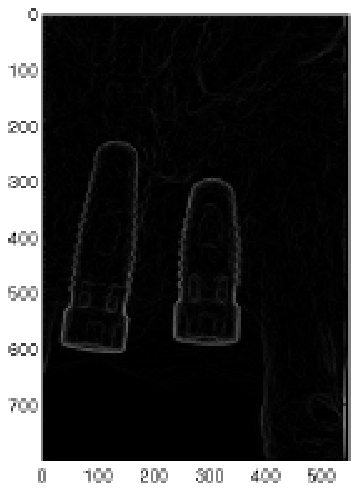}\hspace{2em}
\includegraphics*[width=7.5cm]{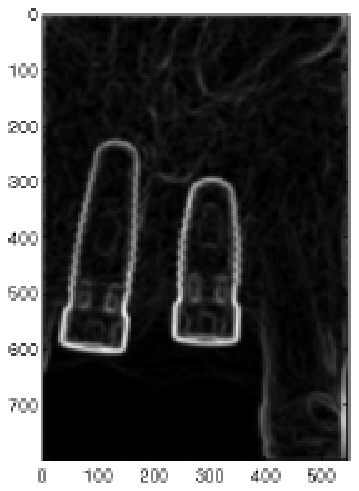}
\end{center}
\caption{\small\it Modulus of the gradient (left) and modulus convolved with a Gaussian kernel (right). \label{IMPLANT_EDGE_EDGE}}
\end{figure}

\begin{figure}
\begin{center}
\includegraphics*[width=7.5cm]{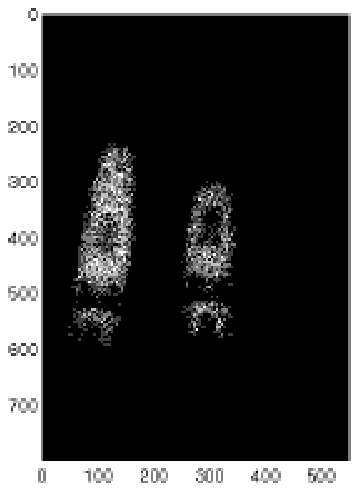}\hspace{2em}
\includegraphics*[width=7.5cm]{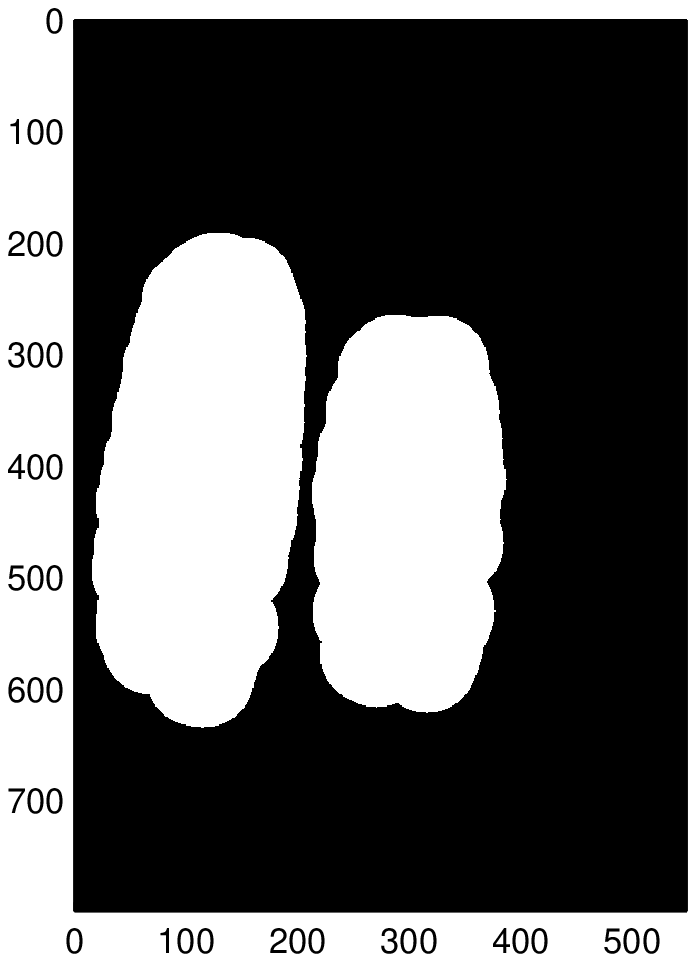}
\end{center}
\caption{\small\it Segmentation using a threshold (left) and patch covering the implants (right). \label{IMPLANT_EDGE_IMPLANT}}
\end{figure}

\begin{figure}
\begin{center}
\includegraphics*[width=7.5cm]{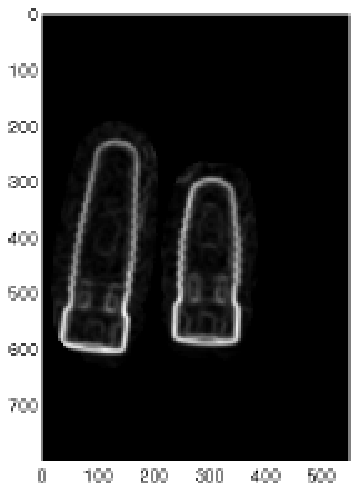}\hspace{2em}
\includegraphics*[width=7.5cm]{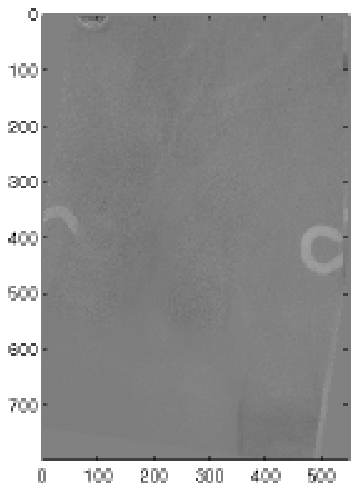}
\end{center}
\caption{\small\it Focus distribution $f$ (left) and subtraction image (right). \label{IMPLANT_EDGE_DIFF}}
\end{figure}

\noindent{\bf Algorithm}
Automatic generation of a focus distribution aiming at the correct registration of a dental implant:
\begin{list}{(\arabic{enumiii}).}{\leftmargin3em \usecounter{enumiii}}
\item
Find (all) edges in the reference image by:
\begin{itemize}
\item median filtering to eliminate ``pepper and salt'' noise from the reference image.
\item computation of the modulus of the gradient (\mbox{Fig. \ref{IMPLANT_EDGE_EDGE} left}).
\item convolution with a Gaussian kernel.
\end{itemize}
This results in an edge distribution focussing all the edges (\mbox{Fig. \ref{IMPLANT_EDGE_EDGE} right}).
\item
Find a patch covering the implant:
\begin{itemize}
\item segmentation using a threshold (\mbox{Fig. \ref{IMPLANT_EDGE_IMPLANT} left}).
\item morphological closing and dilation (\mbox{Fig. \ref{IMPLANT_EDGE_IMPLANT} right}).
\end{itemize}
\item
Create the focus distribution:
\begin{itemize}
\item multiply the patch from step (2) and edge distribution produced in step (1).
\end{itemize}
\end{list}
FMI using this focus distribution results in Fig. \ref{IMPLANT_EDGE_DIFF} right,
demonstrating that both images are well aligned with respect to the dental implant.

\begin{figure}
\begin{center}
\includegraphics*[width=7.5cm]{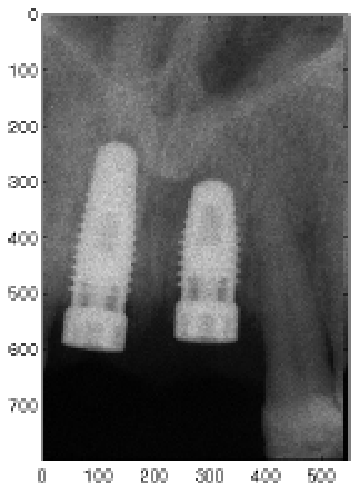}\hspace{2em}
\includegraphics*[width=7.5cm]{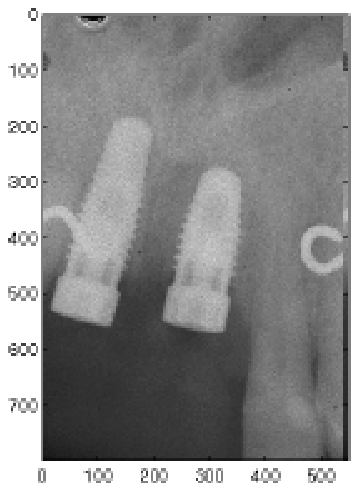}
\end{center}
\caption{\small\it Reference image (left) and Test image (right). \label{IMPLANT_EDGE_Ref_Test}}
\end{figure}

\section{Discussion}\label{Sec:Discussion}
\setcounter{equation}{0}

In this paper we have explored Mutual Information as registration criterion from its information theoretical origin.
The parallelism put forward by Collignon \cite{Collignon1998} between image registration and the model of a communication channel remains unsatisfactory.
The validity of MI cannot be explained from information theory.
Hughes and Daubechies \cite{Hughes2007} identify fundamental properties of MI in the framework of multi-modal image registration,
to introduce simpler alternative similarity measures (distance metric between equivalence classes of images).
Traditional MI neglects spatial information, it is dependent on overlap, and does not allow for the introduction of prior knowledge.
Image registration based on (Normalized) Focussed Mutual Information with respect to a density function is a means to introduce this prior knowledge.
In Jacquet et al.\ \cite{Jacquet2007b} Gauss Focussed Mutual Information is used to eliminate the dependence on overlap.
In Jacquet et al.\ \cite{Jacquet2007} Gauss Focussed Mutual Information is applied to the follow-up of small
dental lesions. The centers of the Gaussian distributions are placed manually by the practitioner on the tooth under
study by means of a digital tool.
In the present paper, several methodologies tailored to specific registration applications are proposed.
In Subsection \ref{subsec:Dentistry}, a dental restoration is detected through segmentation,
and transformed into a regional focus distribution through convolution with a Gaussian kernel.
In Subsection \ref{subsec:Cephalometry}, the purpose of the approach is to create elongated foci along line structures
used as references for the registration.
These lines structures are indicated with a series of points, manually placed by the
practitioner, using a digital tool. B-spline curves are generated using these points
as control points.
The B-splines modeling the skull are transformed into the above mentioned focus distribution by
convolution with a Gaussian kernel.
In Subsection \ref{subsec:Implants}, two methodologies are presented.
In both cases, the interaction between implants and the surrounding bone tissue is studied.
The fact that the implants are simply connected objects in the scene with a maximal radio-opacity constitute the prior knowledge.
Both applications are handled in a fully automated procedure in which the focus is derived from the image
representing the modulus of the gradient.
In the first case the object of the study is the movement of the implant due to aseptic loosening,
which requires focussing on the bone, and therefore, removing the implant from the focus.
In the second case the object of the study is the evolution of the bone tissue
surrounding an implant and therefore, focus is put on the implant.

Further study will combine the efforts to incorporate spatial information with
the introduction of prior knowledge through the sample distribution
and extend it to registration of 3D images, such as hip-, knee-, and shoulder implants.
We will elaborate the automatic creation of models to be used as basis for the sampling distribution.
Alternative functional forms will be studied in order to increase robustness.
Determination of the optimal number of gray value bins is traditionally an aspect of optimal estimation.
We will explore optimal recombination of gray values into gray value bins from a pure registration point of view.
Furthermore, there is no fundamental objection to the use of elastic transformations in combination with FMI registration.
The segmentation technique used in the examples is extremely crude (threshold).
When migrating to e.g.\ elastic transformations of images of soft tissue we will incorporate more subtle image segmentation methods.

\section{Acknowledgements}
\setcounter{equation}{0}

Data lateral radiographies: courtesy of Prof. Guy De Pauw, UGent.

Data bitewing: courtesy of Prof. Peter Bottenberg, UZ Brussel.

Data dental implants: courtesy of Prof. Jan Cosyn, UZ Brussel.

\bibliography{Reg2}
\bibliographystyle{plain}

W. Jacquet  \newline
Vrije Universiteit Brussel, Department of Mathematics - DWIS,
Pleinlaan 2,
B-1050 Brussels,
Belgium,
wjacquet@vub.ac.be

P. de Groen  \newline
Vrije Universiteit Brussel, Department of Mathematics - DWIS,
Pleinlaan 2,
B-1050 Brussels,
Belgium.

\end{document}